# LEARNING MOBILE ROBOT BASED ON ADAPTIVE CONTROLLED MARKOV CHAINS

*V.Ya. Vilisov*
*University of Technology, Energy IT LLP, Russia, Moscow Region, Korolev*
*vvib@yandex.ru*

***Abstract.*** *Herein we suggest a mobile robot-training algorithm that is based on the preference approximation of the decision taker who controls the robot, which in its turn is managed by the Markov chain. Set-up of the model parameters is made on the basis of the data referring to the situations and decisions involving the decision taker. The model that adapts to the decision taker's preferences can be set up either a priori, during the process of the robot's normal operation, or during specially planned testing sessions. Basing on the simulation modelling data of the robot's operation process and on the decision taker's robot control we have set up the model parameters thus illustrating both working capacity of all algorithm components and adaptation effectiveness.*

## Introduction

There are many ways to increase the effectiveness of the robotic system (RS). Thus, we either can provide a high effectiveness of the sensor systems (SS) and/or maximum possible independence of the RS.

Within the framework of the first method, it is possible to increase the informativeness and a number of data collection channels. However, it may also require increasing the productivity of the on-board computer. SS effectiveness can also be increased by using effective algorithms for data processing that are received from them (for example, we can do it by applying different filtration models: Kalman filtration, Bayesian filtration, POMDP etc., as well as using an effective solution of *SLAM* problem etc.). Within the first approach, the task of one of the problems is *to discover the structure and composition of the SS that would possess a minimum sufficiency for the performance of RS's tasks*.

The second approach is closely related to the necessity of providing a special behavior of the RS that would correspond to the objectives and preferences of the decision taker that controls the RS. This direction is closely related to the research undergone in the field of training RS to behave adequately and to interconnect more effectively with the decision taker [1, 2, 3]. Here we make a good use of the Bayesian education tools and Markov processes including POMDP.

The solutions that were obtained when implementing these approaches quite often can complement and/or compensate each other's disadvantages.

The purpose of this work is to solve the problem of selecting the minimum composition of SS that would be sufficient for RS to execute tasks independently. However, due to the limited size of the article, the work concentrates only on teaching independent mobile robots (IMR) preferences of the decision taker and estimating effectiveness of the task performance within the selected configuration of the SS. As an RS we take an IMR possessing two contact (or other functionally similar) sensors and a pair of driving wheels. The task of the robotic system lies in scanning some area (room, territory etc.). Typical example of such IMR is a robotic vacuum cleaner (RVC) or a robotic lawn mower. RS can perform this task by scanning the room using one of the stored programs that is chosen by the decision taker. At that, the effectiveness of the task performance shall vary according to the program selection. It depends on both the program and the room's shape. The logic of the RS is the following: by using different behavior modes applied for similar situations (activation of specific sensors), it makes different decisions. For example, when the left sensor is activated, the RS shall decide to move back taking a right turn etc. In order to simplify the presentation of this work's materials we shall consider the simplest variant of IMR containing two sensors (states) and two decisions (variant actions). In particular, the sensor area is presented by two front (left and right) crash/proximity transducers with the two actions/decisions being the crash reactions (states). The decisions take form of moving back by turning left and moving back by turning right. The increase of the multitude of states and decisions will boost up the variety and effectiveness of IMR behavior not influencing in any way the content of the suggested algorithm. All main mathematical constructions shall be performed for the case with two states and two decisions.

This class of problems is often solved with discrete Markov Decisions Processes (MDP). The suggested algorithm of adapting RS to the objective preferences of the decision taker is founded on the reverse problem for the Markov payoff chain (RPMDP). The problem to be studied here observes effective actions of the decision taker thus computing the estimate of the payoff/objective function of MDP. Therefore, when solving a direct problem for MDP (DPMDP), the optimal controls shall be adapted to the objective preferences of the decision taker.

**Markov Models Applied in RS Management**

*Markov payoff chains* or *Markov profit chains*, which are also called *controlled Markov chains* are the developments of Markov chains, whose description is complemented with the control element, i.e. the decision of the decision taker made when locating Markov chain in one of its states. The decision is presented as a multitude of alternatives and their corresponding multitude of transition probability matrices.

A large group is comprised by the *Partially Observable Markov Decisions Processes (POMDP)*. In comparison with MDP its additional element is the multitude of dimensions. These models contain all components that are inherent to the traditional models of dynamic controlled systems [12], which, in their turn, contain equations of process/system, control and dimensions. This presentation allows us to solve three main groups of management problems: filtration, identification and optimal control.

When using MDP in RS management problems [5, 6, 7, 8, 9], we consider payoff functions (PF) be known and a priori set up when the system or controlling algorithms were developed. Structure and parameters of PF influence the specific value of the optimal decision. If PF does not correspond to the value hierarchy and preference system of the decision taker that controls the RS, the computed solution shall be optimal for the specific PF at the same time being non-optimal for the objective preferences of the decision taker. Therefore, the optimal controlled influence that is obtained by any method is optimal to the PF.

MDP is considered to be set up if we know its such elements as *multitude of states* $i = \overline{1,m}$, *start state probability vector* $\bar{p}_0 = \|p_i\|_m$, *multitude of solutions* $k = \overline{1,K}$, *one-step process transition conditional probability matrix* $P^k = \|p_{ij}^k\|_{mm}^K$, *one-step conditional payoff matrix* $R^k = \|r_{ij}^k\|_{mm}^K$. The solution of MDP is the optimal strategy $f^*$ being one out of many $S$-strategies. The random strategy that possesses $s = \overline{1,S}$ index can be presented as the following column vector: $f^s = [k_1^s \quad k_2^s \quad \cdots \quad k_m^s]^T$. Here $T$ is the conjugation symbol. Hereinafter we shall record column vector as a conjugated row vector for the purposes of compactness. Optimal strategy provides a maximum of one-step accumulated or mean profits/payoffs. Within the strategy structure, $k_i^s$ is the decision that should be made according to $s$-strategy should the process at the current stage $n$ is in state $i$.. The structure of the specific strategy that is applied for usage (decision making) within the current presentation leads to the fact that instead of the multitude of $P^k$ and $R^k$ matrices we shall see the generation of single working matrices, correspondingly $P^s$ and $R^s$.

**Solution of Direct Problem MDP (DPMDP)**

When solving MDP we usually apply [10, 11] a recurrent algorithm based on the principle of Richard Bellman or iteration algorithm of Ronald Howard [11] that allows to improve the solution stepwise. Should the spaces of states and solutions be not large, the optimal solution of the problem may be found with the brute-force search for strategies. We used brute-force search method for the model calculations below.

*Brute-Force Search for Strategies* suggests comparing the competing strategies using the value of *one-step mean payoff* that is existing within the steady-state conditions. The search is performed within the class of stationary strategies. Let us define mean payoff value calculated for one step of $V^s$ for the random $s$-strategy within the steady-state conditions. Let us construct $P^s$ and $R^s$ work matrices for the $s$-strategy. They are formed from the original transition matrices where we use a specific strategy configuration $f^s = [k_1^s \quad k_2^s \quad \cdots \quad k_m^s]^T$ as the key. Thus, the first row of $P^s$ is moved from the first row of $P^{k_1^s}$ matrix, the second row is moved from the second row of $P^{k_2^s}$ matrix and so on. $R^s$ matrix is constructed in the same manner. Thus, for the purposes of $s$-strategy referring to the multitude of $P^k$ and $R^k$ matrices, we can build a single one-step transition matrix $P^s$ and a single one-step payoff matrix $R^s$. Therefore, on condition that the process was in $i$-state, the one-step mean payoff shall be defined as:

$$r_i^s = \sum_{j=1}^{m} p_{ij}^s r_{ij}^s$$

In order to calculate the undoubted mean payoff we shall define the state probability vector within $\bar{f}^N = [p_1^N \quad p_2^N \quad \cdots \quad p_m^N]^T$ steady-state conditions, where $N$ stands for the fact that probabilities correspond to the large step numbers accounting for the steady-state process character. In this case, the mean payoff of $V^s$-step shall be defined as the following value for the $s$-stationary strategy:

$$V^s = \sum_{i=1}^{m} p_i^N r_i^s = \sum_{i=1}^{m} p_i^N \sum_{j=1}^{m} p_{ij}^s r_{ij}^s$$

Should the payoff have a meaning of profit, the optimal strategy selection criterion is:
$$s^* = arg \max_{s \in \{1, S\}} V^s$$
Limited probability vector of states referring to Markov $\bar{p}^N$-process complies with the following matrix equation:
$$(P^s)^T \bar{p}^N = \bar{p}^N$$
At that, the condition of normalization should be performed for the probabilities of states:
$$\sum_{i=1}^{m} p_i^N = 1$$
The solution of the system comprised of two last equations allows obtaining coordinate values of $\bar{p}^N$ vector.

**Solution of Reverse Problem MDP (RPMDP)**

Let us consider what data is included into the observations and what should be found as a result of RPMDP solution. A multitude of presentations is available to the observations. At each $n$-step we can see $i_n$ chain conditions and $k_n$ decisions that are available for observations and which were made by the decision taker, where $i_n, k_n \in \{1; 2\}$. After the end of presentation, the $v$ value of the received payoff shall be computed. Within the context of the example under consideration (IMR scanning robot), the condition is the actuation of the left or right sensor while the decision shall be its moving back making a left or right turn. A useful (without consideration of wastes) duration of the path that was made by the robot during the certain period of time (which is proportional to the scanned area) can serve as each presentation's payoff. Thus, the component of the observation that is considered within the reverse problem solution algorithm is one presentation, i.e. the totality of the alternating conditions, made decisions and presentation's total payoffs.

The result of RPMDP solution is the elements that are necessary for the direct problem (DPMDP) solving, i.e. probability and payoff matrices.

RPMDP solution scheme can be presented with 3 stages.

*1st Stage.* Each presentation contains the $f^s \in \{f^1, f^2, f^3, f^4\}$ strategy, which was chosen by the decision taker. The complete set of strategies is the following: $f^1 = [1 \quad 1]^T$; $f^2 = [1 \quad 2]^T$; $f^3 = [2 \quad 1]^T$; $f^4 = [2 \quad 2]^T$. Referring to each presentation, we estimate the frequencies of some decisions made within some condition. Afterwards, on the basis of these frequencies, we define the closest strategy, which later corresponds to the given presentation.

*2nd Stage.* The whole multitude of transition probability matrices is estimated in relation to each presentation: $P^1, P^2, P^3, P^4$. Each of these matrices is a conditional one (i.e. it is applied when we select the solution that corresponds to the upper index of the matrix). Similar to the 1st stage, we separately compute the frequencies of transitioning from one state to another per each presentation (using the pairs of steps of Markov chain) taking into account the made decision (conditions of the transition). The frequencies that are obtained in this manner are the estimates of one-step MDP conditional transition probabilities. Afterwards these probabilities are put into the corresponding places of the transition probability matrices.

*3rd Stage.* The purpose of constructing the payoff estimates is the following: to calculate the estimates of the payoff vector $r_i^s$ on the basis of the observed parameters (estimates) of the transition probability matrices and according to the payoffs in each observation (presentation). For this we shall use the least square method, whose recurrent form [11] that connects the previous ($q$) estimates of the observations with the current ones, ($q$+1) is the following:
$$\hat{r}_{q+1} = \hat{r}_q + Q_q P_q [P_q^T Q_q P_q + 1]^{-1} [v_{q+1} - P_q^T \hat{r}_q],$$
$$Q_{q+1} = Q_q - Q_q P_q [P_q^T Q_q P_q + 1]^{-1} P_q^T Q_q,$$
where:
$Q_q = (P_q^T P_q)^{-1}$; $v_{q+1}$ is the payoff within the ($q$+1)-numbered observation; $P_q$ is the transition matrices that are obtained on the 2nd stage within $q$-numbered observation.

Within the recurrent estimation equations, the MDP presentation is the *observation step* while the MDP step is actually one step of Markov chain made within the framework of the specific presentation.

With the appearance of each new $q$-numbered presentation of the payoff vector, its estimates are refined recurrently. This is the formal characterization of the decision taker's positive experience made with the help of MDP. Here we can also say that current preferences of the decision taker are approximated by the *Markov chain of decision making*.

The recurrent estimation algorithm not only removes the prior doubt, but also allows adapting to the drift of payoffs, objectives and preferences of the decision taker by correcting the strategies on the basis of the payoff vectors, which are adjusted according to the current observations of the decision taker's actions.

**Model Example**

In order to check the working capacity and effectiveness of the suggested scheme of the reverse problem solution (which is the nucleus of the mobile robot adaptive control procedure) we have conducted the simulation experiment. The given data was formed randomly. One of the parameter variants of the modelled MDP is provided in the table below.

Solving direct problem by the brute-force strategy search showed that the 2nd strategy is the optimal one: $f^2 = [1 \quad 2]^T$. According to its logic, we should choose the first solution for the first process state with the second being chosen for the second one. In this case, the one-step mean payoff shall constitute 71 units within the steady-state mode. Using the Games Theory terminology, this solution corresponds to the decision taker's pure strategy. As a rule, when the operator (decision taker) controls the robot in reality, he/she takes into account a multitude of performance targets (and not only a single one). At that he/she does not "feel" the strategy, whose optimality is based on many criteria, therefore the decision taker may often use his/her own subjective and mixed control strategy.

Table 1. Parameters of Modelled MDP

| Solutions $k$ | Conditions $i$ | One-Step Transition Probabilities $p_{ij}$ | | One-Step Payoffs $r_{ij}$ | |
|---|---|---|---|---|---|
| | | $j=1$ | $j=2$ | $j=1$ | $j=2$ |
| 1 | 1 | **0.05** | **0.95** | **45** | **79** |
| | 2 | **0.19** | **0.81** | **44** | **31** |
| 2 | 1 | **0.27** | **0.73** | **25** | **23** |
| | 2 | **0.48** | **0.52** | **93** | **45** |

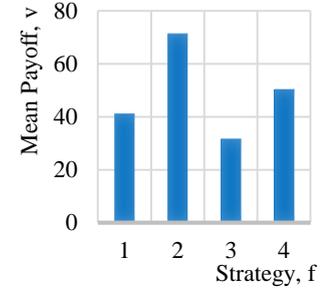

Pic. 1. Mean Payoffs

In order to solve the reverse problem we have simulated 100 presentations, with each of them containing 30 points. It means that the modelled decision taker made 30 decisions in relation to the appearing values of the current states per each presentation. One out of four pure strategies was applied during each presentation. This data was processed in accordance with three stages of the reverse problem solution algorithm that are provided above.

According to the statistics of made decisions, we have definitely identified the pure strategies that were applied by the decision taker at the first stage. This is caused by the fact that each research considered a fully observable MDP.

At the second state we have computed subsequently refined estimations of one-step transition matrices. At that, within the iteration process of the estimate refinement, each of 100 presentations was used as another observation. On Picture 2 you can see step-by-step changes of the estimates referring to 4 probabilities (the total number of matrix probabilities is 8, but 4 of them are independent while the rest 4 are computed as one's complement).

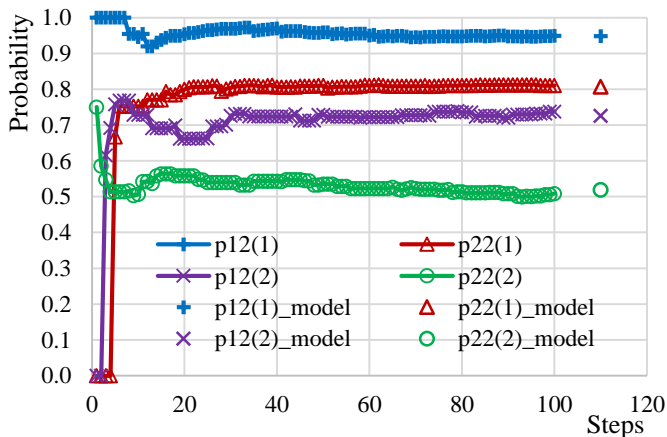

Pic. 2. Convergence of MDP's Probability Estimates

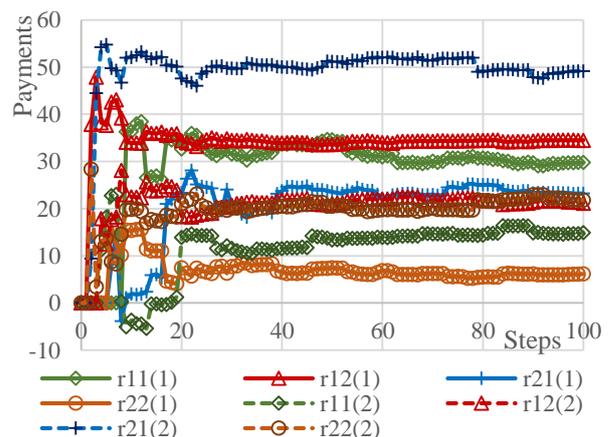

Pic. 3. Convergence of MDP's Payoff Estimates

Modelled probability values (provided in Table 1) are marked separately on Picture 2.

$r_i^k$ estimates of the elements that are folded into the vectors of payoff matrices are computed at the third stage with the help of the observations that are calculated on each step (i.e. that refer to each new presentation) and the payoff that corresponds to the performed presentation according to the recurrent correlations. Their sufficient number for the purposes of the considered dimensionalities is four (similar to the probability estimates). The estimate convergence of these values is shown graphically on Pic. 3, while Pic. 4 contains solutions of the direct problem MDP made on the basis of the step-by-step estimates. It is clearly shown on Pic. 4 that the adaptation process is converging quickly.

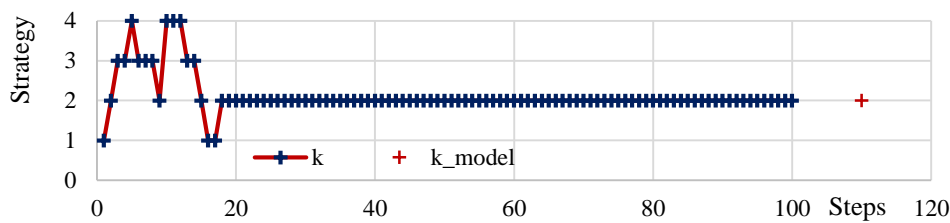

Pic. 4. Convergence of MDP Solutions According to its Parameter Estimates

**Conclusions**

The research that was made for other dimensionalities of state space (capacity of the sensor field) referring to the solution space showed that the solution convergence in relation to the considered class of models stays high. Moreover, the experiment optimal planning tools that are used during the RS education process allow shortening the time that is usually used to adjust the decision taker's preferences.

The preference model that is adjusted and pre-built within the RS is highly adequate to the preferences and objectives of the decision taker, while the quality of decisions that are made by the RS are not inferior to the quality of the decisions made by the "teacher" of MDP-model. Should the signs of the environment non-stationarity appear or should the decision taker's preferences change, the model can be re-adjusted and reloaded into the RS. The process of setting up/educating the model can be made using a separate model or a special testing device, while the MDP-model that was adjusted for new conditions can be loaded as a "hot" update without interrupting normal operation of the RS.

Further development of the suggested approach can unfold in several directions. In particular, we can extend the considered spectrum of the robotic system's sensor field variants and/or use POMDP-models.